\documentclass[letterpaper, 10 pt, conference]{ieeeconf}  

\IEEEoverridecommandlockouts                              

\usepackage{graphicx} 
\usepackage{epsfig} 
\usepackage{algorithmic}
\usepackage{algorithm}
\usepackage{amssymb}
\usepackage{amsmath}
\usepackage{amsfonts}
\usepackage[english]{babel}
\usepackage{setspace,verbatim,longtable,dcolumn,ifpdf,multicol}
\usepackage{cite}
\usepackage{color}
\usepackage{multirow}
\usepackage{url}
\title{\LARGE \bf
Point-Cloud-Based Aerial Fragmentation Analysis for Application in the Minerals Industry
}

\author{Thomas Bamford, Kamran Esmaeili, and Angela P.~Schoellig
\thanks{Thomas Bamford and Kamran Esmaeili are with the Mine Modeling and Analytics Lab (www.omre-researchgroup.com), Lassonde Institute of Mining, University of Toronto, Canada. Email: thomas.bamford@mail.utoronto.ca, kamran.esmaeili@utoronto.ca}%
\thanks{Angela P.~Schoellig is with the Dynamic Systems Lab (www.dynsyslab.org), Institute for Aerospace Studies, University of Toronto, Canada. Email: schoellig@utias.utoronto.ca}%
\thanks{This work was supported by Split-Engineering, the University of Toronto's Dean's Strategic Fund, the Canada Foundation for Innovation John R. Evans Leaders Fund, and the Natural Sciences and Engineering Research Council of Canada.}
}

\begin{document}

\maketitle
\thispagestyle{empty}
\pagestyle{empty}

\begin{abstract}

This work investigates the application of Unmanned Aerial Vehicle (UAV) technology for measurement of rock fragmentation without placement of scale objects in the scene to determine image scale. Commonly practiced image-based rock fragmentation analysis requires a technician to walk to a rock pile, place a scale object of known size in the area of interest, and capture individual 2D images. Our previous work has used UAV technology for the first time to acquire real-time rock fragmentation data and has shown comparable quality of results; however, it still required the (potentially dangerous) placement of scale objects, and continued to make the assumption that the rock pile surface is planar and that the scale objects lie on the surface plane. This work improves our UAV-based approach to enable rock fragmentation measurement without placement of scale objects and without the assumption of planarity. This is achieved by first generating a point cloud of the rock pile from 2D images, taking into account intrinsic and extrinsic camera parameters, and then taking 2D images for fragmentation analysis. This work represents an important step towards automating post-blast rock fragmentation analysis. In experiments, a rock pile with known size distribution was photographed by the UAV with and without using scale objects. For fragmentation analysis without scale objects, a point cloud of the rock pile was generated and used to compute image scale. Comparison of the rock size distributions show that this point-cloud-based method enables producing measurements with better or comparable accuracy (within 10\% of the ground truth) to the manual method with scale objects.

\end{abstract}

\section{INTRODUCTION}
\begin{figure}
\centering
\includegraphics[width=.5\textwidth]{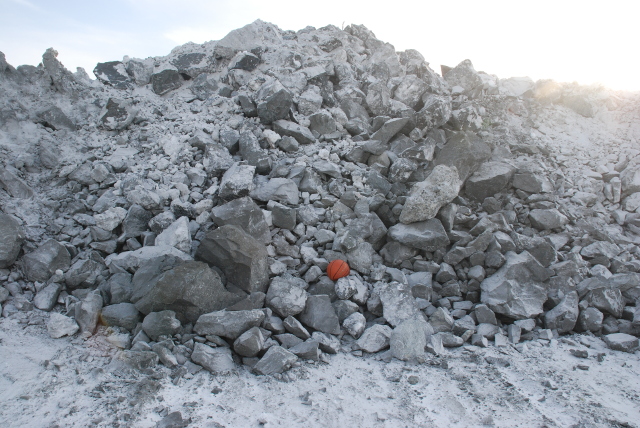}
\caption{Typical photo captured during manual image-based rock fragmentation analysis. The scale object (basket ball) is required in order to determine the size of the rock fragments from single 2D images \cite{esmaieli-mt15}.}
\label{fig:FigureField}
\end{figure}

In recent years, Unmanned Aerial Vehicle (UAV) technology has been introduced to the minerals industry to conduct terrain surveying, monitoring and volume calculation tasks. These tasks are essential for mining operations, but they do not leverage all of the benefits that UAVs can offer to the industry. In general, UAVs can be used for the acquisition of any kind of high-resolution (aerial) data, which may be beneficial in blast design, mill operations and other mine-to-mill optimization campaigns. Moreover, compared to traditional and typically manual measurement techniques, data acquisition with UAVs can be automated to provide higher spatial- and temporal-resolution data, which in turn improves the statistical reliability of measurements. Other benefits of UAV-based data collection in the minerals industry include: no disruption of production, safer for technicians, and being able to collect data from typically inaccessible and hazardous areas. While UAVs can be configured to carry stereo cameras and Light Detection and Ranging (LiDAR) systems, currently commercially available UAVs are configured with high-resolution monocular cameras.

Production blasting in mining operations acts to reduce the size of rock blocks so that the rock can be transported from an in-situ location to downstream mining and comminution processes. Measuring post-blast rock fragmentation is important to many mining operations because rock fragmentation greatly influences the efficiency of all downstream mining and comminution processes. Post-blast rock fragmentation is an important metric for blasting engineers in the optimization of a mining operation. Many methods have been developed throughout human history for estimating rock size distribution, including visual observation by an expert, sieve analysis, and more recently 2D and 3D image analysis~\cite{franklin-ijmge88}.

\begin{figure}
\centering
\includegraphics[width=.5\textwidth]{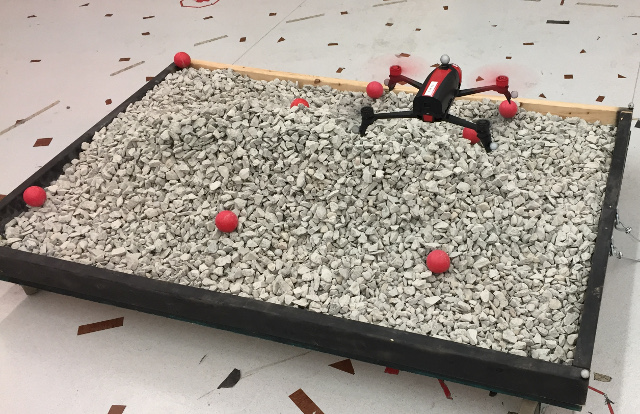}
\caption{Aerial vehicle in lab environment conducting aerial fragmentation analysis with known, red scale objects to determine image scale from single 2D images.}
\label{fig:FigureAction}
\end{figure}

Image analysis techniques for rock fragmentation are commonly used in modern mining operations because they enable practical, fast, and relatively accurate measurement of rock fragmentation~\cite{sanchidrian-rmre09}. There are many established approaches for image analysis for measurement of rock size distribution which use different sensors and approaches for image capturing, processing and data collection~\cite{hunter-mst90}. While more accurate image analysis techniques are being developed using LiDAR sensing~\cite{onederra-mt15} and Deep Neural Network image segmentation~\cite{ramezani-isee17}, the most common technique is to use a monocular camera and scale objects (see Fig.~\ref{fig:FigureField}), and to capture images from fixed ground locations. As identified in our previous work~\cite{bamford-cami16}, there are limitations with the use of fixed monocular cameras and scale objects for rock fragmentation analysis which include:
\begin{itemize}
\item \textit{Limited accuracy:} the surface of the pile is assumed to be planar and the rock size distribution, a 3D quantity, is extracted from the 2D surface (more precisely, images of it).
\item \textit{Low temporal and spatial resolution:} it is time consuming for technicians to conduct it more frequently.
\item \textit{Challenging and potentially hazardous operating environment:} technicians have to work at the base of the rock pile.
\end{itemize}
To overcome these limitations and to automate the data collection process, our work has focused on using UAVs to conduct aerial rock fragmentation analysis (see Fig.~\ref{fig:FigureAction}).

Previous work has been conducted using UAVs with monocular cameras for aerial fragmentation analysis in a controlled lab environment to compare time effort and accuracy with respect to conventional methods~\cite{bamford-cami16} and as a case study in an active quarry~\cite{tamir-isee17}. Both of these works required manual placement of scale objects either physically or virtually (using orthomosaic software), which increases the amount of time required to complete the fragmentation analysis. Also, identified in~\cite{akyildiz-isee17}, the location of the scale object in the image has an impact on the accuracy of the results, because the rock pile is assumed to be planar, which is rarely the case for a rock pile. As a result, the scale objects' locations may result in better or worse planar approximations, making manual placement a potential source of error. Also important to note is that if physical scale objects have to be placed on a rock pile in an active mining environment, the technician is put at risk navigating loose rock and potentially undetonated explosives.

This paper presents a method of computing image scale for aerial fragmentation analysis using a point cloud generated from a series of images and camera poses. The latter are estimated from interoceptive (camera orientation) and global positioning system (GPS) measurements obtained from the onboard UAV sensors. Our work is essential for the automation of aerial fragmentation analysis, because the mining environment rarely provides ideal conditions for placement of scale objects and manual picture-taking.

First, we present a point-cloud-based method used for computing image scale without having to place scale objects and, most importantly, without assuming the pile is perfectly planar. Structure-from-motion (SFM) point cloud creation lies at the core of our proposed methodology. We then introduce the lab environment used for proof-of-concept experiments. We compare our results in terms of accuracy and time effort with the conventional approach (using scale objects). The point-cloud-based method produced rock size distributions with errors less than 10\% relative to ground truth for an average of 5.5 minutes of additional time effort for point cloud generation and image scale computation. In addition, a statistical analysis of repeated experiments shows that aerial fragmentation analysis with and without using scale objects robustly produces the same measurements. Finally, the paper concludes with a discussion of the experimental results and how sources of variability introduced in a field-scale environment may impact accuracy.

\section{METHODOLOGY} \label{sec:method}
This section describes a point-cloud-based method used to calculate image scale. The proposed method uses a monocular camera attached to a UAV, interoceptive sensors onboard the UAV to measure UAV and camera orientation, and global position information (e.g., GPS or an external motion capture system) to measure camera position. First, the UAV flies around the area of interest capturing a series of overlapping images oriented towards the area's center. Then, a point cloud is generated through structure-from-motion (SFM), creating a surface elevation profile in the frame of position measurement, referred to as the world frame (Sec.~\ref{sec:method:sfm}). Next, the UAV captures photos at planned locations throughout the area for fragmentation analysis. At each capture location, the position and orientation of the camera is logged. Next, camera parameters (Sec.~\ref{sec:method:cam}), camera position and orientation (Sec.~\ref{sec:method:pose}) are used to project corner points of the image onto the surface elevation profile. This projection represents the light rays emitted at corner points as line equations in the world frame (Sec.~\ref{sec:method:ray}). These line equations are then intersected with the surface elevation profile to find the locations of the image corners (Sec.~\ref{sec:method:plane}). Finally, these projected points are used to calculate image scale considering the resolution of the image and the distance between corner points in the world (Sec.~\ref{sec:method:scale}).

\subsection{Structure-From-Motion Point Cloud Creation} \label{sec:method:sfm}
\begin{figure}
\centering
\includegraphics[width=.5\textwidth]{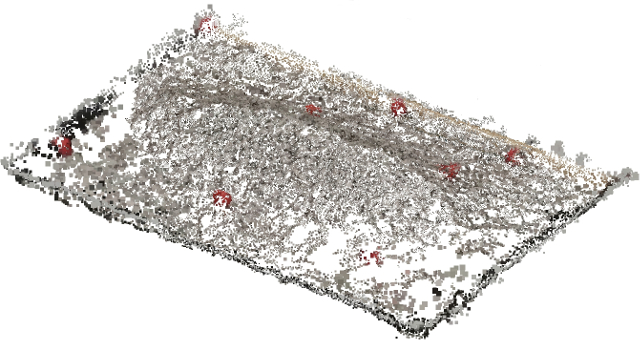}
\caption{The point cloud created for the rock fragment pile with the structure-from-motion algorithm.}
\label{fig:sfm}
\end{figure}

The first step in this process is creating a point cloud. For this step, an open-source structure-from-motion (SFM) software~\cite{wu-vsfm11, wu-cvpr11, wu-sgpu07} is used for sparse and dense 3D reconstruction. The software takes a set of images, detects and describes scale-invariant feature transform (SIFT) features in them, matches these features between images, conducts bundle adjustment to create sparse and then dense 3D reconstruction of the scene, stores these reconstructions as a point cloud, and finally transforms the point cloud using global position information for the set of images. The UAV is used to take overlapping images around the perimeter of the investigation area. For example, if the area of interest is a soccer field, the UAV flies around the perimeter of the field at 2~meters per second capturing photos aimed at the center of the field. The result of using the SFM software for the rock fragment pile in Fig.~\ref{fig:FigureAction} is illustrated in Fig.~\ref{fig:sfm}. When this method is implemented in a large environment, such as a mine, this task can be performed by another UAV prior to or in parallel with a UAV conducting analyses that require the point-cloud-based method for determining image scale. The point cloud created in this step can also be used for other analyses than rock fragmentation analysis, such as drill and blast optimization campaigns~\cite{stewart-isee17}.
\subsection{Camera Parameter Matrix} \label{sec:method:cam}
The next step towards obtaining image scale for fragmentation analysis is the introduction of the camera intrinsic parameters. These parameters are required to transform a point, represented by pixel coordinates, in the image to a point on the image in the world frame. The camera parameter matrix defined as~(cf. \cite{corke-rvc11}):
\begin{equation}
\label{eqn:cam_param}
\textbf{K} = \left[
\begin{IEEEeqnarraybox*}[][c]{,c/c/c,}
f_x & s & c_x\\
0 & f_y & c_y\\
0 & 0 & 1%
\end{IEEEeqnarraybox*}
\right] ,
\end{equation}
where \( f_x \) and \( f_y \) are focal lengths in the x- and y-direction, respectively, \( c_x \) and \( c_y \) are pixel coordinates of the optical center, and \( s \) is the skew between sensor axes. These parameters are innate characteristics of the camera and sensor, and should be estimated through a camera calibration. For the setup in Sec.~\ref{sec:setup}, these parameters are estimated using an open-source camera calibration package~\cite{bowman-ros17}. 
\subsection{Camera Pose} \label{sec:method:pose}
\begin{figure}
\centering
\includegraphics[width=.5\textwidth]{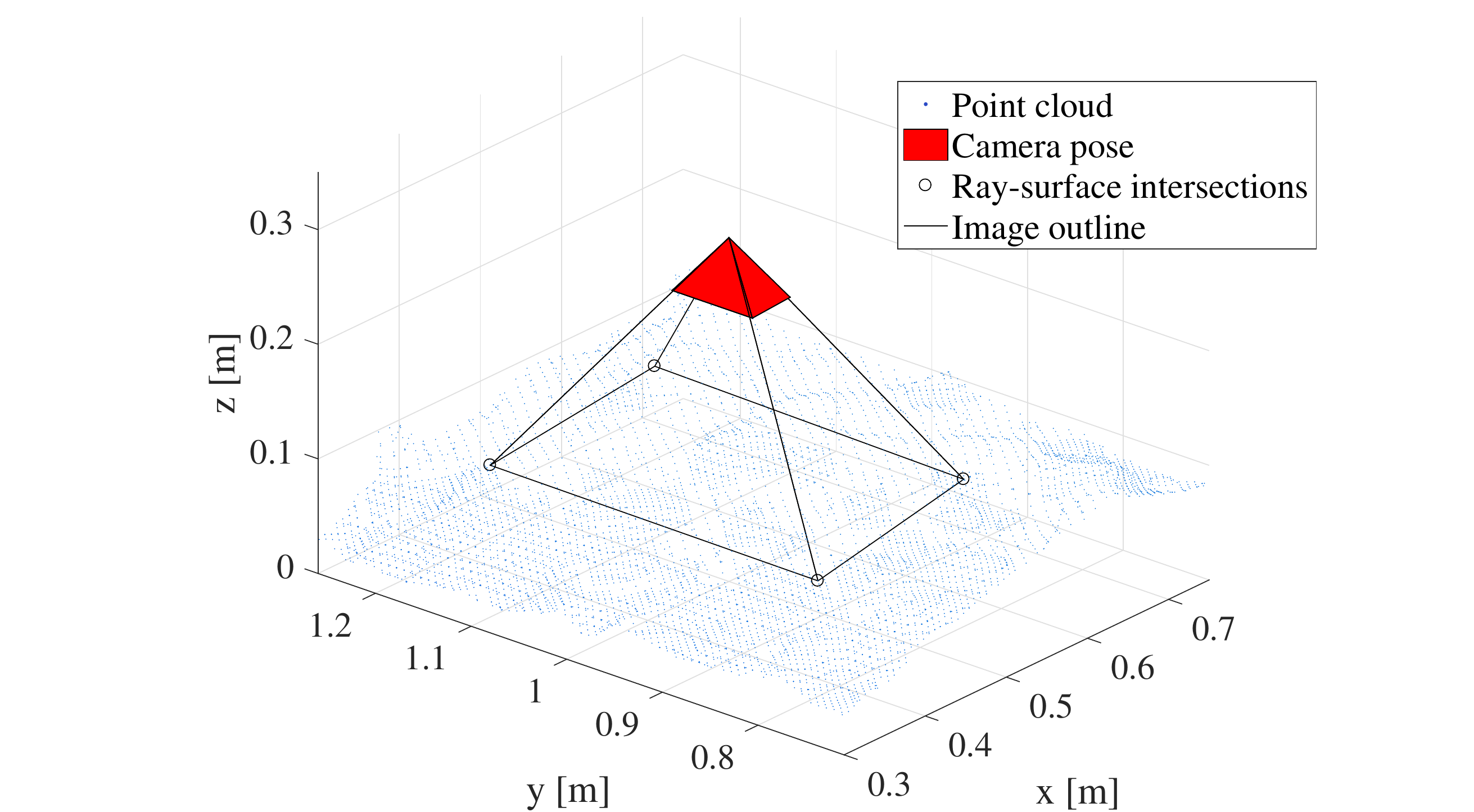}
\caption{Camera pose above the rock pile in the world frame. The image outline is computed by projecting rays from the four corners of the image to intersect with the rock pile surface.}
\label{fig:cloud}
\end{figure}

The pose, translation and rotation, of the camera in the world frame is required as an origin in the derivation of a ray equation needed to project image points in the world frame and onto the surface elevation profile. The pose of the camera is defined as:
\begin{equation}
\label{eqn:cam_extrinsic}
\textbf{T} = \bigg[
\begin{IEEEeqnarraybox*}[][c]{,c/c,}
\textbf{C} & \textbf{r}   \\
\textbf{0}^T & 1 %
\end{IEEEeqnarraybox*}
\bigg] ,
\end{equation}
where \( \textbf{T} \) is referred to as the (\( 4\times4 \)) transformation matrix of the camera with respect to the world frame origin, \( \textbf{C} \) is a (\( 3\times3 \)) rotation matrix in the special orthogonal group,  \textit{SO}(3), and \( \textbf{r} = (x_{r}, y_{r}, z_{r}) \) is a translation vector. This transformation matrix is known as the camera's extrinsic parameters, and comprises a minimum of six parameters to describe translation and rotation in the special Euclidean group,  \textit{SE}(3). For our experiments in Sec.~\ref{sec:setup}, the pose of the camera is estimated from onboard measurement of the camera orientation and the UAV pose is obtained from a motion capture system. In field experiments, the UAV pose will be estimated by fusing odometry and GPS sensor measurements. Fig.~\ref{fig:cloud} illustrates the camera pose above the rock pile, represented by a point cloud.
\subsection{Ray Equation}  \label{sec:method:ray}
This section derives an equation to represent a ray from pixel coordinates in the image to the world frame using the camera parameter matrix (Sec.~\ref{sec:method:cam}) and the camera pose (Sec.~\ref{sec:method:pose}). The derived equation is used to represent each ray projected from the four corners of the image so that the rays can be intersected with the surface elevation profile. Fig.~\ref{fig:cloud} illustrates the four rays projected from the image corners to intersect with the point cloud. A pixel location in the image, \(\textbf{p}\), is represented by coordinates \(u\)  and \(v\), where \(u\)  and \(v\) are integers. This pixel location, \(\textbf{p}\), is represented in the image using homogeneous coordinates, \( \tilde{\textbf{p}} = (u',v',w') \). The non-homogeneous pixel coordinates are computed from:
\begin{equation}
\label{eqn:nonhomogeneous}
u = \dfrac{u'}{w'} ,\quad v = \dfrac{v'}{w'} .
\end{equation}
To represent the point, \( \textbf{p} \), in the image as \( \tilde{\textbf{p}} \), we set \( w'=1 \) such that that the center of the image frame is the origin and points are mapped to the plane \( w'=1 \). The first step to find an equation to represent a ray from image pixel coordinates to the world frame, is to determine the direction of the ray in the image. The direction of the ray in the image, \( \tilde{\textbf{P}}_{c} \), from the homogeneous pixel location, \(\tilde{\textbf{p}}\), for a camera with a camera parameter matrix from (\ref{eqn:cam_param}), \( \textbf{K} \), is calculated using:
\begin{equation}
\label{eqn:ray_image}
\tilde{\textbf{p}}_{c} = \textbf{K}^{-1}\tilde{\textbf{p}} .
\end{equation}
Using \( \tilde{\textbf{p}}_{c} \) from~(\ref{eqn:ray_image}), the ray direction in the image is transformed into the world frame. This transformation assumes that camera distortion has been removed from the image to ensure that the ray in the world frame is straight and follows the same direction as the ray in the image frame. In the setup described in Sec.~\ref{sec:setup}, the camera distortion is removed using an open-source image processing package~\cite{mihelich-ros17} with the camera distortion model parameters estimated during the camera calibration. The homogeneous ray direction (\(4 \times 1 \)) in the world frame, \( \tilde{\textbf{p}}_{m} \), is computed using the ray direction in the image, \( \tilde{\textbf{p}}_{c} \), and the camera pose from (\ref{eqn:cam_extrinsic}), \(\textbf{T} \), according to:
\begin{equation}
\label{eqn:ray_world}
\tilde{\textbf{p}}_{m} = \textbf{T}\left[
\begin{IEEEeqnarraybox*}[][c]{,c,}
\tilde{\textbf{p}}_{c} \\
1 %
\end{IEEEeqnarraybox*}
\right] ,
\end{equation}
where \( \tilde{\textbf{p}}_{m}=(x_{m},y_{m},z_{m},1) \), with \( x_{m},y_{m},z_{m}\) representing the slope along each axis in the world frame. Once the ray direction in the world frame, \( \tilde{\textbf{p}}_{m} \), is determined using~(\ref{eqn:ray_world}), the equation for the ray in the world frame emitting from the pixel location, \(\textbf{p}\), is represented by:
\begin{equation}
\label{eqn:ray}
\tilde{\textbf{q}} = \left[
\begin{IEEEeqnarraybox*}[][c]{,c,}
\textbf{r} \\
1 %
\end{IEEEeqnarraybox*}
\right] + \alpha\tilde{\textbf{p}}_{m} ,
\end{equation}
where \( \tilde{\textbf{q}}=(x,y,z,1) \) is a homogeneous point on the ray at the location \( \textbf{q}=(x,y,z) \) in the world frame, \( x \), \( y \), and \( z \) are components in the world frame, and \( \alpha \) is a scalar since the pixel location is projected along the ray. For example, if (\ref{eqn:ray}) is used to represent the ray emitted from a point in the image in the world frame, and assuming that the ground surface lies on a plane at \( z = 0 \) then (\ref{eqn:ray}) can be easily rearranged to solve for \( \alpha \):
\begin{equation}
\label{eqn:ray_z0}
\alpha = -\dfrac{z_{r}}{z_{m}} .
\end{equation}
\subsection{Plane and Line Parameterization} \label{sec:method:plane}
To find the intersection of the ray with the point cloud, the point cloud is triangulated using the Delaunay triangulation and each triangle is represented as a plane. The three corners of each triangle (\( \textbf{p}_{0}, \textbf{p}_{1}, \textbf{p}_{2} \)) represent a plane such that a general point on the plane is represented by:
\begin{equation}
\label{eqn:general_plane}
\textbf{p}_{0} + (\textbf{p}_{1} - \textbf{p}_{0})\eta + (\textbf{p}_{2} - \textbf{p}_{0})\mu ,
\end{equation}
with \( \eta,\mu\in\mathbb{R} \). Using two points along the corner point ray, such as the translation vector of the camera pose \( \textbf{p}_{a} = \textbf{r} \) and the point intersecting the plane at \( z = 0 \), \( \textbf{p}_{b} \), a simple line equation is developed:
\begin{equation}
\label{eqn:general_line}
\textbf{p}_{a} + (\textbf{p}_{b} - \textbf{p}_{a})t ,
\end{equation}
with \( t\in\mathbb{R} \). The line and plane parameters at the point of intersection can then be solved according to: \newpage
\begin{equation}
\label{eqn:intersection}
\left[
\begin{IEEEeqnarraybox*}[][c]{,c,}
t \\
\eta \\
\mu %
\end{IEEEeqnarraybox*}
\right] = \left[
\begin{IEEEeqnarraybox*}[][c]{,c,c,c,}
(x_a-x_b) & (x_1-x_0) & (x_2-x_0) \\
(y_a-y_b) & (y_1-y_0) & (y_2-y_0) \\
(z_a-z_b) & (z_1-z_0) & (z_2-z_0) %
\end{IEEEeqnarraybox*}
\right]^{-1} \left[
\begin{IEEEeqnarraybox*}[][c]{,c,}
(x_a-x_0) \\
(y_a-y_0) \\
(z_a-z_0) %
\end{IEEEeqnarraybox*}
\right] .
\end{equation}
\subsection{Image Scale for Fragmentation Analysis} \label{sec:method:scale}
Once all four corner points of the image are represented by~(\ref{eqn:ray}), the scale is calculated. For the specialized fragmentation analysis software used in Sec.~\ref{sec:setup}, the scale is applied at the top and bottom edge of the image. As such, each pair of corner points along each edge are used to compute the image scale:
\begin{equation}
\label{eqn:scale}
\text{scale} = \dfrac{\text{image width}}{\sqrt{(\Delta x)^2 + (\Delta y)^2 + (\Delta z)^2}} ,
\end{equation}
where the image width is the width in pixels, and \( \Delta x, \Delta y,\) and \( \Delta z \) are the distances between corner points along the \(x, y \) and \(z \) world frame axes, respectively. The distance between the corner points is in the unit of distance measurement used in the image analysis software. Assumptions made in the image software for rock fragmentation analysis are described in Sec.~\ref{sec:setup:split}.
\subsection{Algorithm} \label{sec:method:algorithm}
\begin{figure}
\centering
\includegraphics[width=.5\textwidth]{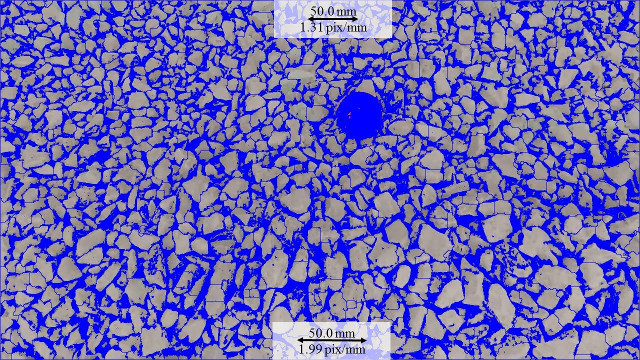}
\caption{Delineated photo in automated aerial fragmentation analysis. Scales computed using the point-cloud-method are shown at the top and bottom of the image. Scale variation from left to right could also be computed, the software Split-Desktop can only set top and bottom scales.}
\label{fig:delineationPCScale}
\end{figure}

Algorithm~\ref{alg} is used to compute scale for each image captured in the aerial fragmentation analysis. Fig.~\ref{fig:cloud} illustrates graphically the corner point intersections found for a given camera pose using this algorithm. Fig.~\ref{fig:delineationPCScale} illustrates the scales computed for the raw photo in Fig.~\ref{fig:raw} using the developed point-cloud-based algorithm.
\begin{algorithm}
\caption{Calculate image scale using the point-cloud-based method.}
\label{alg}
\begin{algorithmic}
\STATE Store all corner point rays defined by (\ref{eqn:ray}) as lines for (\ref{eqn:general_line})
\FOR {each triangle created to represent the point cloud}
\STATE parameterize 3 points as a plane using (\ref{eqn:general_plane})
\FOR {each corner point line}
\STATE compute intersection using (\ref{eqn:intersection})
\IF {point on ray, (\( t \geq 0 \)), and inside triangle ( \( \eta,\mu \in[0,1] \) and \( \eta+\mu \leq 1 \))}
\STATE store intersection point for use in (\ref{eqn:scale})
\ENDIF
\ENDFOR
\ENDFOR
\IF {intersection point not found}
\STATE assume on plane \( z = 0 \)
\ENDIF
\STATE compute image scale using intersections with (\ref{eqn:scale})
\end{algorithmic}
\end{algorithm}

\section{EXPERIMENTAL SETUP} \label{sec:setup}
A lab experiment was designed and set up to demonstrate the feasibility and benefits of automated aerial rock fragmentation analysis. This step was deemed necessary before conducting any tests in large-scale field experiments because in the laboratory we have a ground truth rock size distribution and ideal conditions for vehicle flight. Fig.~\ref{fig:FigureAction} illustrates the UAV in flight conducting aerial fragmentation analysis in the lab environment. The hardware choices, lab configuration and procedure used to conduct image analysis are presented in the following subsections. More details are available in~\cite{bamford-cami16}.

\subsection{Rock Fragment Pile}
A pile of rock fragments with different sizes, ranging from coarse gravel (19~millimeters) to fine sand (\(<\)4~millimeters), was built in the lab. Prior to forming the pile, the rock fragments were put through sieve analysis to determine the `true' rock size distribution as a reference for experimental results. The results of the sieve analysis are presented for four discrete screen sizes, which is referred to as the discrete sieve series in Fig.~\ref{fig:distribution}. To use the sieve analysis as a reference, a rock size distribution curve was fit to the collected data. The parameters of this distribution are found in~\cite{bamford-cami16}. Spherical scale objects, with a diameter of 60 millimeters, were used to provide image scale when applying conventional image analysis, as seen in Fig.~\ref{fig:FigureAction}. These scale objects are ignored and masked when applying the point-cloud-based method for image scale computation.


\subsection{Lab Environment}
The indoor lab is equipped with a motion capture system for precise UAV localization and control. The lab has fluorescent lighting, providing optimal lighting conditions for the image analysis. The environment is also free of wind.

\subsection{Unmanned Aerial Vehicle and Software Framework}
A commercially available UAV with integrated camera, the Parrot Bebop 2, was used in our experiments. This UAV has the ability to capture stabilized high-resolution photos and videos, which is essential for accurate image analysis. In this experiment, the UAV broadcasts a video stream with an image resolution of \( 1280 \times 720 \) pixels which is stabilized onboard with respect to the world frame during flight. The camera orientation is changed onboard by moving a virtual window through the field of view of the integrated fish-eye-lens. The UAV receives camera commands and transmits the camera orientation in tilt and pan with respect to the world frame.

The open-source Robot Operating System (ROS)~\cite{ros-icra09} was chosen to act as the central software node of the experimental setup. In these experiments, ROS uses a predetermined high-level flight plan and actual position and orientation measurements from the motion capture system to send low-level velocity and camera control commands wirelessly to the UAV. Images are captured from the UAV video stream and then analyzed in the specialized fragmentation analysis software. We use a macro to run the analysis automatically.

\subsection{Rock Fragmentation Analysis} \label{sec:setup:split}
\begin{figure}
\centering
\includegraphics[width=.5\textwidth]{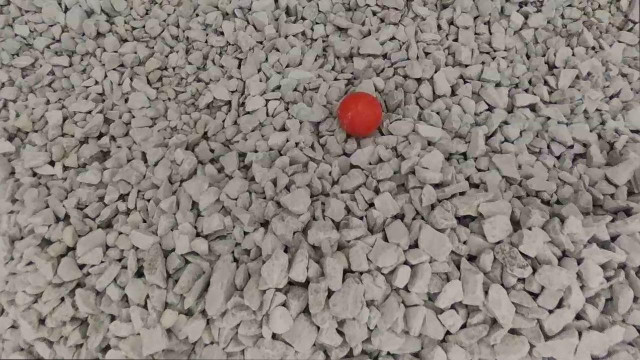}
\caption{Raw photo captured in automated aerial fragmentation analysis with scale object to determine image scale.}
\label{fig:raw}
\end{figure}

\begin{figure}
\centering
\includegraphics[width=.5\textwidth]{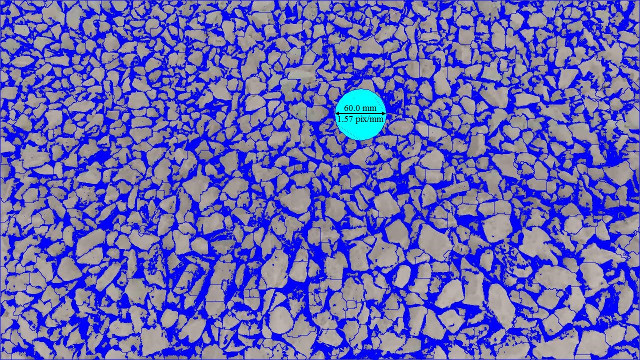}
\caption{Delineated photo in automated aerial fragmentation analysis. Scale object is measured and masked in light blue.}
\label{fig:delineationScale}
\end{figure}

For these experiments, Split-Desktop~\cite{split-10}, an industry standard software for image analysis in mining, was used. The main software parameters, such as the fines factor, were calibrated using sieve analysis data as a reference. The software receives an image and delineates particles using image segmentation, see Fig.~\ref{fig:delineationScale}. A scale object is then traced graphically to set the image scale assuming that the spherical scale object lies on the rock pile surface and that the surface is planar. Optionally, an image scale can be set uniformly or at the top and bottom edge of the image without graphical input assuming that the scale changes linearly from top to bottom. Fig.~\ref{fig:raw} gives an example of a raw photo imported into Split-Desktop, and Fig.~\ref{fig:delineationScale} illustrates the same photo after image segmentation.

\subsection{Flight Plan} \label{sec:setup:flight}
\begin{figure}
\centering
\includegraphics[width=.5\textwidth]{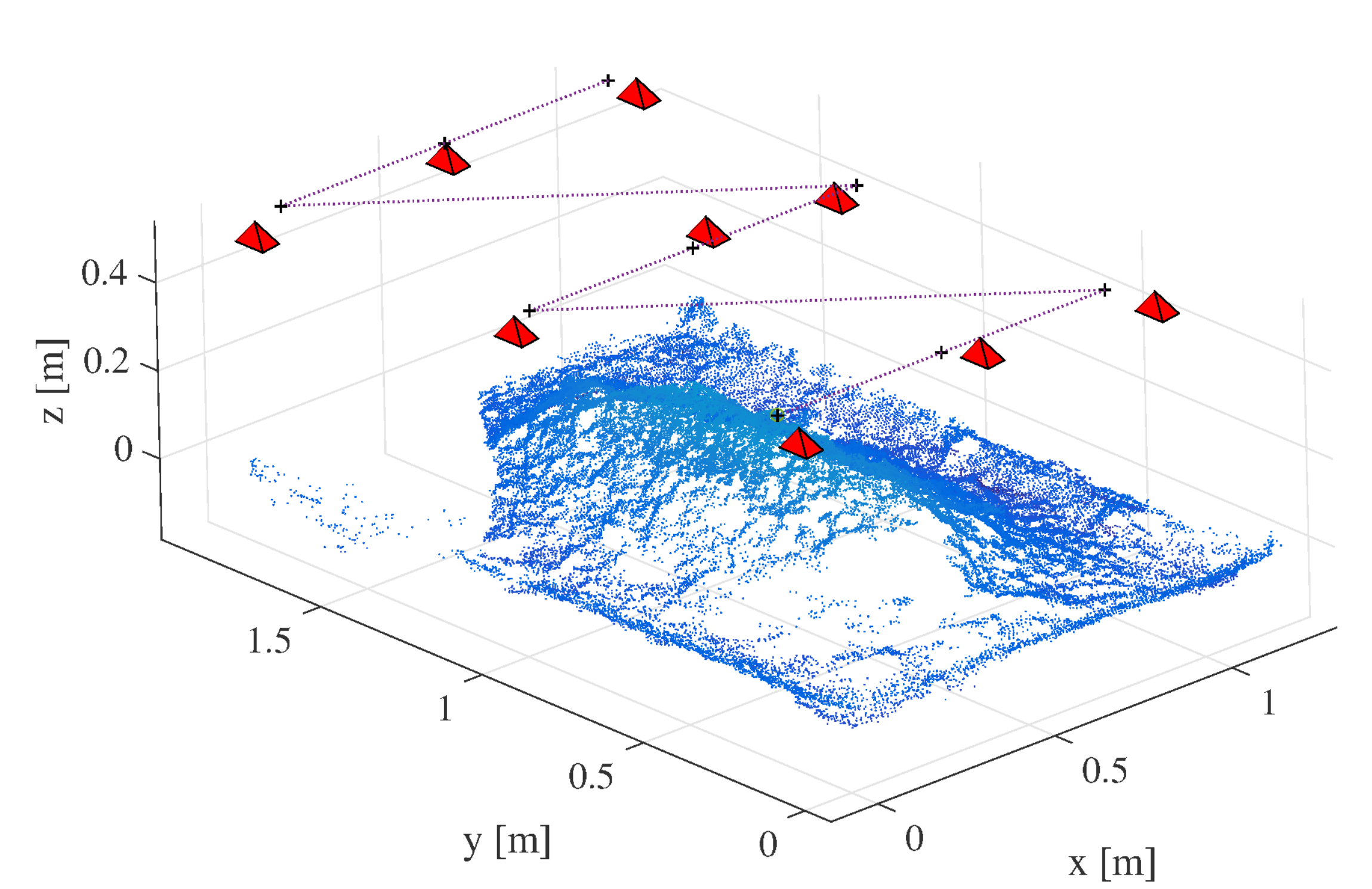}
\caption{Planned flight over the rock pile. Camera poses (in red) are included for a sample trial. Crosses indicate planned image capture locations and the dotted line represents the flight trajectory.}
\label{fig:flight}
\end{figure}

For these experiments, a flight plan was created to capture photos for a tilt angle of 83~degrees, at a fixed altitude of 0.5~meters above the rock pile base while ensuring no image overlap. The tilt angle was chosen so that the camera was directed as far downward as the UAV specification allowed such that the images are approximately perpendicular to the rock pile surface. This is following the suggestions made in the Split-Desktop software. In future work, adjusting the camera angle according to the pile geometry will be investigated. Fig.~\ref{fig:flight} illustrates the flight plan over the rock pile with planned and actual image capture locations for a sample trial. Each planned  UAV location captures a single scale object near the center of the photo. This is a fair comparison to conditions in the mine environment since measurement devices are sparsely placed on a rock pile for rock fragmentation analysis campaigns such that the largest area possible can be captured.

\section{EXPERIMENTAL RESULTS}
The following section presents experimental results of applying the method described in Sec.~\ref{sec:method} (point-cloud-based method) in terms of accuracy and time effort compared with placing scale objects for image scale (scale-object method). It then presents a statistical analysis of repeated experiments to illustrate the robustness of aerial fragmentation analysis in the scale-object and point-cloud method cases.

\subsection{Rock Size Distribution}
Using the experimental setup described in Sec.~\ref{sec:setup}, ten trials of automated aerial fragmentation were conducted and a rock size distribution was generated for the rock pile. The flight plan and rock pile morphology remained constant for all ten trials. Nine photos per trial were taken by the UAV at the planned locations according to the flight plan described in Sec.~\ref{sec:setup:flight} to calculate a rock size distribution. Fig.~\ref{fig:raw} illustrates one of these photos for a sample trial with a single scale object near the center of the image. For each trial, fragmentation analysis was conducted on the same set of photos using both the scale-object method and the point-cloud-based method. For each method, the same delineation net (a parameter in Split-Desktop) was used and masking was applied to the scale object and pile boundaries. Figures~\ref{fig:delineationPCScale} and \ref{fig:delineationScale} show an example of a scaled and delineated photo using the scale-object and the point-cloud-based method, respectively. The following subsections provide a comparison between the scale-object and point-cloud-based method in terms of prediction accuracy and time effort for a sample trial.

\subsubsection{Prediction Accuracy}
\begin{figure*}
\centering
\includegraphics[width=0.97\textwidth]{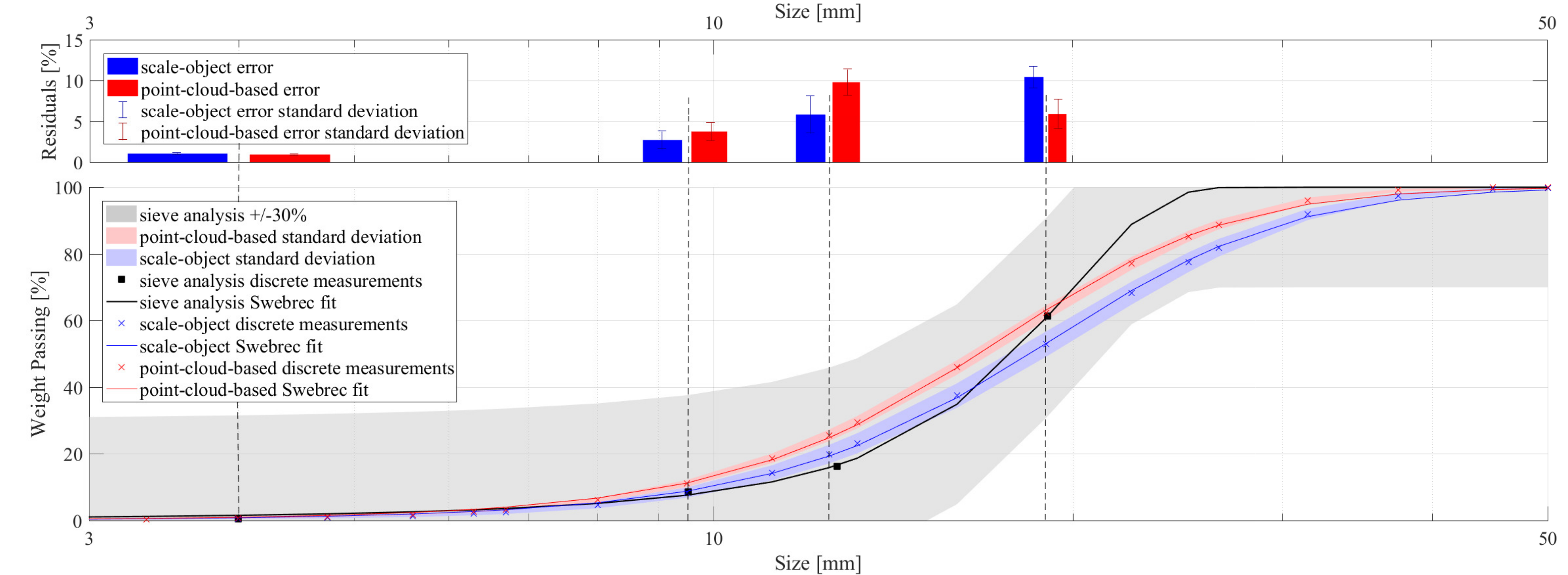}
\caption{Automated aerial rock fragmentation analysis results for ten trials using the point-cloud-based and scale-object methods with respect to the sieve analysis reference curve (ground truth). Discrete points (average value) and standard deviation envelopes represent the combined results for all ten trials. The Swebrec rock size distribution function~\cite{sanchidrian-irfb15} has been fit to the discrete points from sieve analysis, scale-object method, and point-cloud-based method so that a 2-norm error between the two image analysis methods and the ground truth could be calculated. The gray envelope represents the accepted maximum error envelope of 30\% recommended by~\cite{sanchidrian-rmre09} for industry standard 2D image analysis of measuring rock fragmentation.}
\label{fig:distribution}
\end{figure*}

To determine prediction accuracy of each method for comparison, the percent error residuals for percent passing with respect to the reference sieve analysis curve were computed for the discrete sieve series. The method of computing percent error residuals is described in detail in~\cite{bamford-cami16}. The average rock size distribution for ten repeated trials with residuals for each method is plotted in Fig.~\ref{fig:distribution}. For this plot, the point-cloud-based method is shown to have comparable accuracy to the scale-object method. The 2-norm error was calculated over the full curve for both methods and the point-cloud-based method has a 6\% improvement over the scale-object method. The point-cloud-based method performs better in the coarse region of the rock size distribution but slightly over-predicts the amount of fines. Both methods can be seen to have residuals less than 10\% with rock size distributions remaining within the accepted maximum error envelope of 30\% recommended by~\cite{sanchidrian-rmre09} for industry standard 2D image analysis of measuring rock fragmentation. All of the other trials exhibited similar trends, as shown by the standard deviation envelopes in Fig.~\ref{fig:distribution} where the point-cloud-based method has a smaller envelope than the scale-object method. The decrease in standard deviation realized by the point-cloud-based method is thought to be caused by a better estimation of scale throughout the image rather than assuming that the pile is planar. This is because actual photo locations varied in the trials, causing the scale location in the images to change, while the point-cloud-based method accounted for this change, the scale-object method did not. These results are very promising since the point-cloud-based method has comparable accuracy and lies well within the industry accepted bounds, which makes it a suitable replacement for the scale-object method during field experiments.

\subsubsection{Time Effort}
Table~\ref{table:times} details the amount of time taken in seconds for the sampling flight and each extra task required for the point-cloud-based method for each trial. For the first trial, the total time taken in addition to flight time and fragmentation analysis in Split-Desktop was 5.7 minutes. Obviously, the point-cloud-based method requires more time effort than the scale-object method in the lab environment due to generating the point cloud and since the rock pile is easily accessible and only covers a small area. However, in field experiments, the amount of time for scale object placement is expected to be much longer, while the amount of time required for the point-cloud-based method is expected to increase marginally. Additionally, most of the time taken was in the construction of a point cloud, which can be used in other analyses for the mining operation.

\begin{table*}[!htb]
\centering
\caption{Trial times for scale-object and point-cloud-based methods in seconds.}\label{table:times}
\begin{tabular}{|c|c|c|c|c|c|c|c|}\hline
Trial & SFM Flight & Sampling Flight & Computing Matches & Sparse Recst. & Dense Recst. & Scale Comp. & Total \\
\hline
1 & 93 & 159 & 121 & 14 & 81 & 30 & 339 \\
2 & 93 & 162 &  88 & 12 & 54 & 27 & 274 \\
3 & 89 & 160 & 124 & 27 & 64 & 29 & 333 \\
4 & 87 & 165 & 129 & 19 & 74 & 33 & 342 \\
5 & 85 & 157 & 130 & 16 & 59 & 34 & 323 \\
6 & 94 & 151 & 119 & 18 & 80 & 34 & 345 \\
7 & 88 & 156 & 124 & 18 & 56 & 32 & 318 \\
8 & 86 & 157 & 121 & 14 & 69 & 29 & 319 \\
9 & 90 & 158 & 184 & 20 & 49 & 7 & 349 \\
10 & 86 & 148 & 191 & 39 & 63 & 30 & 409 \\
\hline
Average & 89 & 157 & 133 & 20 & 65 & 29 & 335 \\
\hline
\end{tabular}
\end{table*}

\subsection{Analysis of Variance}
The Analysis of Variance (ANOVA) is a statistical model used to analyze whether there are any statistically significant differences between the means of independent factors. One-dimensional ANOVA with replication was set up to analyze whether repeated aerial fragmentation analysis statistically produces the same rock size distribution. This analysis was set up based on other applications of ANOVA for comparing rock size distributions, more details on the assumptions made and the background of ANOVA is available in~\cite{niedoba-ms16}. The rock size distributions for all ten trials were used to conduct one-dimensional ANOVA with replication for the point-cloud-based and scale-object methods, respectively. The trial and weight percent passing are sources of variability. However, we are interested in the effect of varying trial because we are testing whether the aerial fragmentation analysis is robust.

The critical value \( F_{n,m;\alpha} \) of the Fisher-Snedecor distribution was set as \( F_{9,20;0.05} = 2.39 \), to have a 5\% level of significance. This critical value is used to reject the null hypothesis that results are the same with varied trials if the \(F \)-test statistic is greater than \( 2.39 \). Tables~\ref{table:anovaPC} and \ref{table:anova} show the \( F \)-test statistic computed for all trials and for the full sieve series for both point-cloud-based and scale-object methods, respectively. As can be seen, both \( F \)-test statistics are much less than \( 2.39 \), and the F-test statistic for the point-cloud-based method is less than the scale-object method. Therefore, the results for all ten trials are statistically the same with a 5\% level of significance for both the point-cloud-based method and the scale-object method. This indicates that aerial fragmentation analysis with and without scale objects is robust and statistically produces the same results when experiments are replicated.

\begin{table}[!htb]
\centering
\caption{Results of one-way ANOVA of replicated experiments for point-cloud-based method.} \label{table:anovaPC}
\begin{tabular}{!{\vrule width 2pt}p{1.3cm}!{\vrule width 2pt}p{1.3cm}!{\vrule width 2pt}p{1.0cm}!{\vrule width 2pt}p{1.0cm}!{\vrule width 2pt}p{1.0cm}!{\vrule width 2pt}}\hline
Source of Variation & Degrees of Freedom & Sum of Squares & Mean Square & \textit{F}-test \\
\hline 
Trial & 9 & 1.9 & 0.21 & 0.003 \\
\hline
Residuals & 20 & 1255.3 & 62.76 & \\
\hline
\end{tabular}
\end{table}

\begin{table}[!htb]
\centering
\caption{Results of one-way ANOVA of replicated experiments for scale-object method.}\label{table:anova}
\begin{tabular}{!{\vrule width 2pt}p{1.3cm}!{\vrule width 2pt}p{1.3cm}!{\vrule width 2pt}p{1.0cm}!{\vrule width 2pt}p{1.0cm}!{\vrule width 2pt}p{1.0cm}!{\vrule width 2pt}}\hline
Source of Variation & Degrees of Freedom & Sum of Squares & Mean Square & \textit{F}-test \\
\hline
Trial & 9 & 8.6 & 0.96 & 0.016 \\
\hline
Residuals & 20 & 1170.0 & 58.50 & \\
\hline
\end{tabular}
\end{table}

\addtolength{\textheight}{-0.9cm}   

\section{DISCUSSION}
Having shown that the point-cloud-based method produces comparable accuracy to the scale-object method, we will apply this method to aerial fragmentation analysis in field-scale experiments. The results have shown that using the point-cloud-based method reduced variability while providing a reduction in error. This is thought to be caused by applying image scale in a more representative manner than assuming that a uniform scale exists or that the rock pile is planar. It is important to note that the results are as only good as the inputs used to reach them. In the laboratory-scale environment, the motion capture system provides UAV localization with 2~millimeter accuracy which has aided the precision of the point-cloud-based method. In an outdoors environment, when localization must rely on noisier GPS sensor measurements, we may expect that the point cloud and point-cloud-based computation of image scale will not be as accurate and that a decrease in repeatability will be the result. Nevertheless, the proposed point-cloud-based method is considered a valuable tool for automating aerial rock fragmentation analysis with UAV technology.

\section{CONCLUSIONS}
This paper proposed a method to calculate image scale for point-cloud-based aerial fragmentation as an alternative for scale objects. We showed that it is equally accurate compared to conventional image scaling suggesting that no scale objects are needed for aerial fragmentation analysis. This will make the process faster, safer and more reliable when applied in the mining environment. Through statistical analysis of replicated experiments, this paper also shows that our aerial fragmentation analysis for both the conventional and point-cloud-based scale computation is robust and produces results that are statistically the same with a 5\% level of significance. The main benefit of using UAVs for aerial fragmentation analysis is that data can be acquired fast and often over a large area, which improves the overall reliability of image-based fragmentation analysis and reduces sampling error. With the proposed method for computing scale using point cloud information, this work takes a step towards the full automation of aerial fragmentation analysis.
%




\bibliographystyle{IEEEtran}

\begin{thebibliography}{10}
\providecommand{\url}[1]{#1}
\csname url@samestyle\endcsname
\providecommand{\newblock}{\relax}
\providecommand{\bibinfo}[2]{#2}
\providecommand{\BIBentrySTDinterwordspacing}{\spaceskip=0pt\relax}
\providecommand{\BIBentryALTinterwordstretchfactor}{4}
\providecommand{\BIBentryALTinterwordspacing}{\spaceskip=\fontdimen2\font plus
\BIBentryALTinterwordstretchfactor\fontdimen3\font minus
  \fontdimen4\font\relax}
\providecommand{\BIBforeignlanguage}[2]{{%
\expandafter\ifx\csname l@#1\endcsname\relax
\typeout{** WARNING: IEEEtran.bst: No hyphenation pattern has been}%
\typeout{** loaded for the language `#1'. Using the pattern for}%
\typeout{** the default language instead.}%
\else
\language=\csname l@#1\endcsname
\fi
#2}}
\providecommand{\BIBdecl}{\relax}
\BIBdecl

\bibitem{esmaieli-mt15}
K.~Esmaieli and J.~Hadjigeorgiou, ``Application of {DFN–-DEM} modelling in
  addressing ground control issues at an underground mine,'' \emph{Mining
  Technology}, vol. 124, no.~3, pp. 138--149, 2015.

\bibitem{franklin-ijmge88}
J.~A. Franklin, N.~H. Maerz, and C.~P. Bennett, ``Rock mass characterization
  using photoanalysis,'' \emph{International Journal of Mining and Geological
  Engineering}, vol.~6, no.~2, pp. 97--112, 1988.

\bibitem{sanchidrian-rmre09}
J.~A. Sanchidri\'{a}n, P.~Segarra, F.~Ouchterlony, and L.~M. L\'{o}pez, ``On
  the accuracy of fragment size measurement by image analysis in combination
  with some distribution functions,'' \emph{Rock Mechanics and Rock
  Engineering}, vol.~42, no.~1, pp. 95--116, 2009.

\bibitem{hunter-mst90}
G.~C. Hunter, C.~McDermott, N.~J. Miles, A.~Singh, and M.~J. Scoble, ``A review
  of image analysis techniques for measuring blast fragmentation,''
  \emph{Mining Science and Technology}, vol.~11, no.~1, pp. 19--36, 1990.

\bibitem{onederra-mt15}
I.~Onederra, M.~J. Thurley, and A.~Catalan, ``Measuring blast fragmentation at
  {Esperanza} mine using high-resolution {3D} laser scanning,'' \emph{Mining
  Technology}, vol. 124, no.~1, pp. 34--36, 2015.

\bibitem{ramezani-isee17}
M.~Ramezani, S.~Nouranian, I.~Bell, B.~Sameti, and S.~Tafazoli, ``Fast rock
  segmentation using artificial intelligence to approach human-level
  accuracy,'' in \emph{{Proc. of the 43rd Annual Conference on Explosives and
  Blasting Technique}}, 2017.

\bibitem{bamford-cami16}
\BIBentryALTinterwordspacing
T.~Bamford, K.~Esmaeili, and A.~P. Schoellig, ``A real-time analysis of rock
  fragmentation using {UAV} technology,'' in \emph{{Proc. of the International
  Conference on Computer Applications in the Minerals Industries (CAMI)}},
  2016. [Online]. Available: \url{http://arxiv.org/abs/1607.04243}
\BIBentrySTDinterwordspacing

\bibitem{tamir-isee17}
R.~Tamir, ``Utilization of aerial drones to optimize blast and stockpile
  fragmentation,'' in \emph{{Proc. of the 43rd Annual Conference on Explosives
  and Blasting Technique}}, 2017.

\bibitem{akyildiz-isee17}
\"{O}zge Aky{\i}ld{\i}z and T.~H\"{u}daverdi, ``Some issues in the image
  analysis application for bench face characteristics and blast fragmentation
  measurement,'' in \emph{{Proc. of the 43rd Annual Conference on Explosives
  and Blasting Technique}}, 2017.

\bibitem{wu-vsfm11}
\BIBentryALTinterwordspacing
C.~Wu. (2011) {VisualSFM}: A visual structure from motion system. [Online].
  Available: \url{http://ccwu.me/vsfm/}
\BIBentrySTDinterwordspacing

\bibitem{wu-cvpr11}
C.~Wu, S.~Agarwal, B.~Curless, and S.~M. Seitz, ``Multicore bundle
  adjustment,'' in \emph{Proc. of the Conference on Computer Vision and Pattern
  Recognition (CVPR)}, 2011.

\bibitem{wu-sgpu07}
\BIBentryALTinterwordspacing
C.~Wu. (2007) {SiftGPU}: A {GPU} implementation of scale invariant feature
  transform ({SIFT}). [Online]. Available:
  \url{http://cs.unc.edu/~ccwu/siftgpu}
\BIBentrySTDinterwordspacing

\bibitem{stewart-isee17}
D.~Stewart and T.~Wiseman, ``Drill and blast improvement project using
  photogrammetry, {GPS} and drill navigation systems,'' in \emph{{Proc. of the
  43rd Annual Conference on Explosives and Blasting Technique}}, 2017.

\bibitem{corke-rvc11}
P.~Corke, \emph{Robotics, Vision and Control}.\hskip 1em plus 0.5em minus
  0.4em\relax Springer-Verlag Berlin Heidelberg, 2011.

\bibitem{bowman-ros17}
\BIBentryALTinterwordspacing
J.~Bowman and P.~Mihelich. (2017) {Robot Operating System (ROS)} camera
  calibration package. [Online]. Available:
  \url{http://wiki.ros.org/camera_calibration}
\BIBentrySTDinterwordspacing

\bibitem{mihelich-ros17}
\BIBentryALTinterwordspacing
P.~Mihelich, K.~Konolige, and J.~Leibs. (2017) {Robot Operating System (ROS)}
  image processing package. [Online]. Available:
  \url{http://wiki.ros.org/image_proc}
\BIBentrySTDinterwordspacing

\bibitem{ros-icra09}
M.~Quigley, K.~Conley, B.~P. Gerkey, J.~Faust, T.~Foote, J.~Leibs, R.~Wheeler,
  and A.~Y. Ng, ``{ROS}: an open-source {Robot Operating System},'' in
  \emph{Proc. of the IEEE International Conference on Robotics and Automation
  (ICRA) Workshop on Open Source Software}, 2009.

\bibitem{split-10}
\BIBentryALTinterwordspacing
{Split Engineering LLC.} (2010) {Split-Desktop}: Fragmentation analysis
  software. [Online]. Available:
  \url{https://www.spliteng.com/products/split-desktop-software/}
\BIBentrySTDinterwordspacing

\bibitem{sanchidrian-irfb15}
J.~A. Sanchidri\'{a}n, ``Ranges of validity of some distribution functions for
  blast-fragmented rock,'' in \emph{{Proc. of the 11th International Symposium
  on Rock Fragmentation by Blasting (FRAGBLAST)}}, 2015.

\bibitem{niedoba-ms16}
T.~Niedoba and P.~Pi\c{e}ta, ``Applications of {ANOVA} in mineral processing,''
  \emph{Mining Science}, vol.~23, pp. 43--54, 2016.

\end{thebibliography}


\end{document}